%% file: main.tex
\documentclass[final]{cvpr}

\usepackage{times}
\usepackage{epsfig}
\usepackage{graphicx}
\usepackage{amsmath}
\usepackage{amssymb}

\usepackage[usenames,dvipsnames]{color}
\usepackage{makecell}
\usepackage{cuted}
\usepackage[misc]{ifsym}
\usepackage{balance}
\usepackage{tabularx}
\newcolumntype{Y}{>{\centering\arraybackslash}X}
\usepackage{amsmath,bm}

\definecolor{citecolor}{RGB}{65,105,225}
\usepackage[pagebackref=true,breaklinks=true,letterpaper=true,colorlinks,citecolor=citecolor,bookmarks=false]{hyperref}

\newcommand\blfootnote[1]{%
  \begingroup
  \renewcommand\thefootnote{}\footnote{#1}%
  \addtocounter{footnote}{-1}%
  \endgroup
}

\def\cvprPaperID{821} % *** Enter the CVPR Paper ID here
\def\confYear{CVPR 2021}

\begin{document}

%%%%%%%%% TITLE
\title{DeeperForensics Challenge 2020 on Real-World Face Forgery Detection:\\[3pt]
Methods and Results}

\author{
Liming Jiang, Zhengkui Guo, Wayne Wu, Zhaoyang Liu, Ziwei Liu, Chen Change Loy,\\[2pt]
Shuo Yang, Yuanjun Xiong, Wei Xia, Baoying Chen, Peiyu Zhuang, Sili Li, Shen Chen, Taiping Yao,\\[2pt]
Shouhong Ding, Jilin Li, Feiyue Huang, Liujuan Cao, Rongrong Ji, Changlei Lu, Ganchao Tan
}

\maketitle

\input{sections/abstract.tex}
\input{sections/introduction.tex}
\input{sections/challenge_summary.tex}
\input{sections/solutions.tex}
\input{sections/discussion.tex}

{\small
\bibliographystyle{ieee_fullname}
\bibliography{sections/egbib}
}

\end{document}

%% file: sections/abstract.tex
% !TEX root = ../main.tex

\begin{abstract}
\label{sec:abstract}

This paper reports methods and results in the DeeperForensics Challenge 2020 on real-world face forgery detection.
The challenge employs the DeeperForensics-1.0 dataset, one of the most extensive publicly available real-world face forgery detection datasets, with 60,000 videos constituted by a total of 17.6 million frames.
The model evaluation is conducted online on a high-quality hidden test set with multiple sources and diverse distortions.
A total of 115 participants registered for the competition, and 25 teams made valid submissions.
We will summarize the winning solutions and present some discussions on potential research directions. 
\blfootnote{
\vspace{-0.3cm}
\begin{itemize}
	\item Liming Jiang, Ziwei Liu, and Chen Change Loy are with S-Lab, Nanyang Technological University.
	\vspace{-0.1cm}
	\item Zhengkui Guo is with The Chinese University of Hong Kong.
	\vspace{-0.1cm}
	\item Wayne Wu and Zhaoyang Liu are with SenseTime Research.
	\vspace{-0.1cm}
	\item Shuo Yang, Yuanjun Xiong, and Wei Xia are with Amazon Web Services.
	\vspace{-0.1cm}
	\item Baoying Chen, Peiyu Zhuang, and Sili Li are with Shenzhen Key Laboratory of Media Information Content Security, Shenzhen University.
	\vspace{-0.1cm}
	\item Shen Chen, Liujuan Cao, and Rongrong Ji are with Media Analytics and Computing Lab, Xiamen University.
	\vspace{-0.1cm}
	\item Taiping Yao, Shouhong Ding, Jilin Li, and Feiyue Huang are with YouTu Lab, Tencent, Shanghai.
	\vspace{-0.1cm}
	\item Changlei Lu and Ganchao Tan are with University of Science and Technology of China.
\end{itemize}
}

\end{abstract}

%% file: sections/introduction.tex
% !TEX root = ../main.tex

\section{Introduction}
\label{sec:introduction}

Recent years have witnessed exciting progress~\cite{DFL, DFLPaper, DeepFakes, faceswap-GAN, faceshifter, deeperforensics1} in automatic face swapping.
Indeed, these techniques have eschewed the cumbersome hand-crafted face manipulation processes, hence facilitating the development of various popular softwares for face editing.
From another perspective, these easy-to-access softwares, named ``Deepfakes'', have also brought risks for being misused and spread. Tampered videos on the internet could lead to possible perilous consequences, entailing the potential legitimate concerns among the general public and authorities.
Therefore, effective face forgery detection methods become an urgent need to safeguard against these photorealistic fake videos, particularly in real-world scenarios where the video sources and distortions are unknown.

We organize the DeeperForensics Challenge 2020 with the aim to advance the state-of-the-art in face forgery detection. Participants are expected to develop robust and generic methods for forgery detection in real-world scenarios.
The challenge uses DeeperForensics-1.0~\cite{deeperforensics1}, a large-scale real-world face forgery detection dataset that contains $60,000$ videos with a total of $17.6$ million frames\footnote{\hspace{0.07cm}Project page: \href{https://liming-jiang.com/projects/DrF1/DrF1.html}{https://liming-jiang.com/projects/DrF1/DrF1.html}.}.
All source videos in DeeperForensics-1.0 are carefully collected, and fake videos are generated by a newly proposed end-to-end face swapping framework. Extensive real-world perturbations are applied to obtain a more challenging benchmark of larger scale and higher diversity.
The dataset also features a hidden test set, which is richer in distribution than the publicly available training set, suggesting a better setting to simulate real-world scenarios.
Besides, the hidden test set will be continuously updated to get future versions along with the development of Deepfakes technology.
The evaluation of the challenge is performed online on the current version of the hidden test set.

In the following sections, we will describe the DeeperForensics Challenge 2020, summarize the winning solutions and results, and provide discussions to take a closer look at the current status and possible future development of real-world face forgery detection.

%% file: sections/challenge_summary.tex
% !TEX root = ../main.tex

\section{About the Challenge}
\label{sec:challengesummary}

\subsection{Platform}
\label{sec:platform}

The DeeperForensics Challenge 2020 is hosted on the CodaLab platform\footnote{\hspace{0.07cm}Challenge website: \href{https://competitions.codalab.org/competitions/25228}{https://competitions.codalab.org/competitions/25228}.} in conjunction with ECCV 2020, The 2nd Workshop on Sensing, Understanding and Synthesizing Humans\footnote{\hspace{0.07cm}Workshop website: \href{https://sense-human.github.io}{https://sense-human.github.io}.}. The online evaluation is conducted using Amazon Web Services (AWS)\footnote{\hspace{0.07cm}Online evaluation website: \href{https://aws.amazon.com}{https://aws.amazon.com}.}.
First, participants register their teams on the CodaLab challenge website. Then, they are requested to submit their models to the AWS evaluation server (with one 16 GB Tesla V100 GPU for each team) to perform the online evaluation on the hidden test set. When the evaluation is done, participants receive the encrypted prediction files through an automatic email. Finally, they submit the result file to the CodaLab challenge website.

\subsection{Dataset}
\label{sec:dataset}

The DeeperForensics Challenge 2020 employs the DeeperForensics-1.0 dataset~\cite{deeperforensics1} that was proposed in CVPR 2020. DeeperForensics-1.0 contains $60,000$ videos constituted by a total of $17.6$ million frames.
The dataset features three appealing properties: good quality, large scale, and high diversity.

To ensure good quality, extensive data collection is conducted. The high-resolution ($1920\times1080$) source videos are collected from $100$ paid actors with four typical skin tones across $26$ countries. Their eight expressions (\ie, neutral, angry, happy, sad, surprise, contempt, disgust, fear) are recorded under nine lighting conditions by seven cameras at different locations. We further ask the actors to perform $53$ supplementary expressions defined by 3DMM blendshapes~\cite{3dmm} to make the dataset more diverse. Besides, a robust end-to-end face swapping framework, DF-VAE, is developed to generate the fake videos. In addition, seven types of real-world perturbations at five intensity levels are applied to obtain a more challenging benchmark of larger scale and higher diversity. Readers are referred to~\cite{deeperforensics1} for details. 

An indispensable component of DeeperForensics-1.0 is the hidden test set, which is richer in distribution than the publicly available training set. The hidden test set suggests a better real-world face forgery detection setting:
\textbf{1)} Multiple sources. Fake videos in-the-wild should be manipulated by different unknown methods;
\textbf{2)} High quality. Threatening fake videos should have high quality to deceive human eyes;
\textbf{3)} Diverse distortions. Different perturbations should be considered.
The hidden test set will evolve by including more challenging samples along with the development of Deepfakes technology.
The evaluation of the challenge is performed on its current version.

All the participants using the DeeperForensics-1.0 dataset should agree to its Terms of Use~\cite{termsdeeperforensics1}.
They are recommended but not restricted to train their algorithms on DeeperForensics-1.0.
The use of any external datasets should be disclosed and follow the Terms of Use.

%DrF1 info (number), cite terms of use, FF++, hidden (info), external data (follow their terms of use)

\subsection{Evaluation Metric}
\label{sec:evaluationmetric}

Similar to Deepfake Detection Challenge (DFDC)~\cite{DFDCWeb}, the DeeperForensics Challenge 2020 uses the binary cross-entropy loss (BCELoss) to evaluate the performance of face forgery detection models:

\begin{footnotesize}
\vspace{-0.35cm}
\begin{equation*}
\small
\label{eq:11}
    \mathrm{BCELoss}=-\frac{1}{N}\sum_{i=1}^N{\left[y_i\cdot\log{\left(p\left(y_i\right)\right)}+(1-y_i)\cdot\log{\left(1-p\left(y_i\right)\right)}\right]},
\end{equation*}
\vspace{-0.2cm}
\end{footnotesize}

\noindent
where $N$ is the number of videos in the hidden test set, $y_i$ denotes the ground truth label of video $i$ (fake: $1$, real: $0$), and $p\left(y_i\right)$ indicates the predicted probability that video $i$ is fake.
A smaller BCELoss score is better, which directly contributes to a higher ranking. If the BCELoss score is the same, the one with less runtime will achieve a higher ranking. To avoid an infinite BCELoss that is both too confident and wrong, the score is bounded by a threshold value.

\subsection{Timeline}
\label{sec:timeline}

The DeeperForensics Challenge 2020 lasted for nine weeks -- eight weeks for the \textit{development phase} and one week for the \textit{final test phase}.

The challenge officially started at the ECCV 2020 SenseHuman Workshop on August 28, 2020, and it immediately entered the development phase. In the development phase, the evaluation is performed on the \textit{test-dev} hidden test set, which contains $1,000$ videos representing general circumstances of the full hidden test set. The \textit{test-dev} hidden test set is used to maintain a public leaderboard. Participants can conduct four online evaluations (each with 2.5 hours of runtime limit) per week.

The final test phase started on October 24, 2020. The evaluation is conducted on the \textit{test-final} hidden test set, containing $3,000$ videos (also including test-dev videos) with a similar distribution as test-dev, for the final competition results. A total of two online evaluations (each with 7.5 hours of runtime limit) are allowed.
The final test phase ended on October 31, 2020.

Finally, the challenge results were announced in December 2020. In total, $115$ participants registered for the competition, and $25$ teams made valid submissions.

%start, phases, final results December, 2020, evaluation chances/runtime limit/video number diff phases, teams

%% file: sections/solutions.tex
% !TEX root = ../main.tex

\section{Results and Solutions}
\label{sec:solutions}

%\subsection{Final Results}
%\label{sec:results}

\begin{table}[h]
\vspace{-0.1cm}
\centering
\footnotesize
\caption{Final results of the top-5 teams in the DeeperForensics Challenge 2020. The runtime is shown in seconds.}
\vspace{0.1cm}
\begin{tabularx}{\linewidth}{c|*{4}{|Y}}
\Xhline{1pt}
Ranking& TeamName& UserName& BCELoss$\downarrow$& Runtime$\downarrow$ \\
\cline{1-5}
1& Forensics& BokingChen& 0.2674& 7690 \\
2& RealFace& Iverson& 0.3699& 11368 \\
3& VISG& zz110& 0.4060& 11012 \\
4& jiashangplus& jiashangplus& 0.4064& 16389 \\
5& Miao& miaotao& 0.4132& 19823 \\
\Xhline{1pt}
\end{tabularx}
\label{tbl:results}
%\vspace{-0.6cm}
\end{table}

Among the $25$ teams who made valid submissions, many participants achieve promising results. We show the final results of the top-5 teams in Table~\ref{tbl:results}. In the following sections, we will present the winning solutions of top-3 entries.

\subsection{Solution of First Place}
\label{sec:solution1}

\noindent
\textit{Team members: Baoying Chen, Peiyu Zhuang, Sili Li}

\begin{figure}[h]
	\centering
%	\vspace{-0.35cm}
	\includegraphics[width=\linewidth]{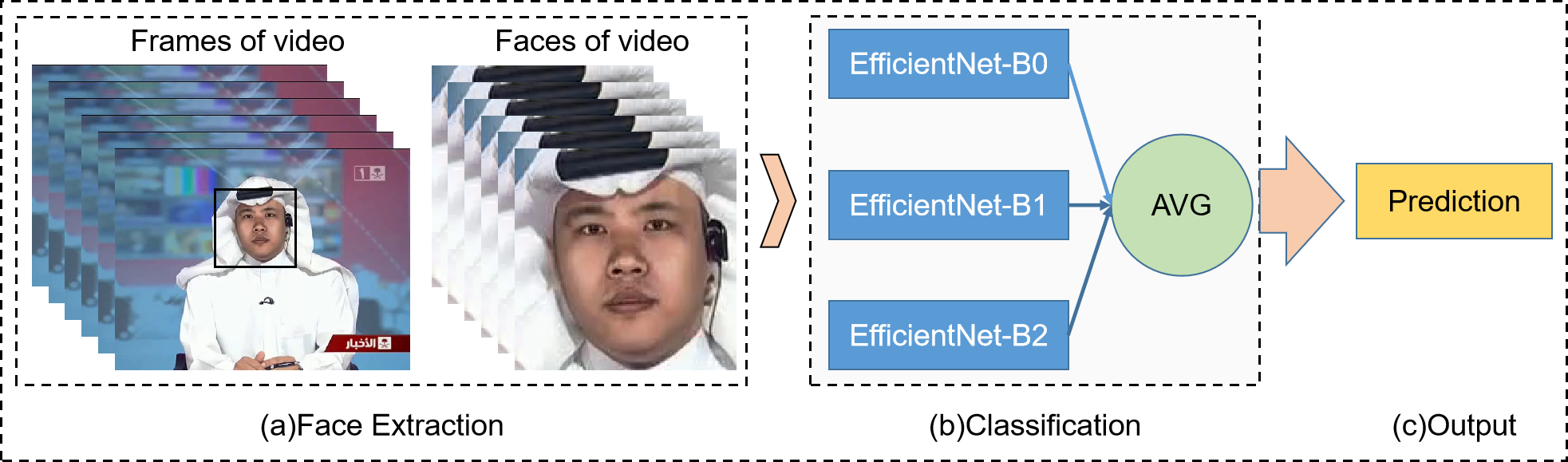}
%	\vspace{-0.5cm}
	\caption{The framework of the first-place solution.}
	\label{fig:solution1}
%	\vspace{-0.16cm}
\end{figure}

As shown in Figure~\ref{fig:solution1}, the method designed by the champion team contains three stages, namely Face Extraction, Classification, and Output.

\noindent
\textbf{Face Extraction.}
They first extract $15$ frames from each video at equal intervals using VideoCapture of OpenCV. Then, they use the face detector MTCNN~\cite{mtcnn} to detect the face region of each frame and expand the region by $1.2$ times to crop the face image.

\noindent
\textbf{Classification.}
They define the prediction of the probability that the face is fake as the face score. They use EfficientNet~\cite{efficientnet} as the backbone,  which was proven effective in the Deepfake Detection Challenge (DFDC)~\cite{DFDCWeb}. The results of three models (EfficientNet-B0, EfficientNet-B1 and EfficientNet-B2) are ensembled for each face. 

\noindent
\textbf{Output.}
The final output score of a video is the predicted probability that the video is fake, which is calculated by the average of face scores for the extracted frames.

\noindent
\textbf{Implementation Details.}
The team employs EfficientNet pre-trained on ImageNet as the backbone. They select EfficientNet-B0, EfficientNet-B1, and EfficientNet-B2 for the model ensemble. In addition to DeeperForensics-1.0, they use some other public datasets, \ie, UADFV~\cite{UADFV}, Deep Fake Detection~\cite{google}, FaceForensics++~\cite{FF++iccv}, Celeb-DF~\cite{celebdfcvpr}, and DFDC Preview~\cite{DFDC}. They balance the class samples with the down-sampling mode. The code of the champion solution has been made publicly available\footnote{\hspace{0.07cm}\href{https://github.com/beibuwandeluori/DeeperForensicsChallengeSolution}{https://github.com/beibuwandeluori/DeeperForensicsChallengeSolution}.}.

\noindent
\textit{$\bullet$ Training:}
Inspired by the DFDC winning solution, appropriate data augmentation could contribute to better results. As for the data augmentation, the champion team uses the perturbation implementation in DeeperForensics-1.0~\cite{drf1_aug} during training. They only apply the image-level distortions: color saturation change (CS), color contrast change (CC), local block-wise (BW), white Gaussian noise in color components (GNC), Gaussian blur (GB) and JPEG compression (JPEG). They randomly mixup these distortions with a probability of 0.2. Besides, they also try other data augmentation~\cite{dfdc_aug}, but the performance improvement is slim.
The images are resized to $224\times224$. The batch size is $128$, and the total training epoch is $50$. They use AdamW optimizer~\cite{adamw} with initial learning rate of 0.001. Label smoothing is applied with a smoothing factor of $0.05$.

\noindent
\textit{$\bullet$ Testing:}
The testing pipeline follows the three stages in Figure~\ref{fig:solution1}. They clip the prediction score of each video in a range of $[0.01, 0.99]$ to reduce the large loss caused by the prediction errors.
In addition to the best BCELoss score, their fastest execution speed may be attributed to the use of the faster face extractor MTCNN and the ensemble of three image-level models with fewer parameters.

\subsection{Solution of Second Place}
\label{sec:solution2}

\noindent
\textit{Team members: Shen Chen, Taiping Yao, Shouhong Ding, Jilin Li, Feiyue Huang, Liujuan Cao, Rongrong Ji}

\begin{figure}[h]
	\centering
%	\vspace{-0.35cm}
	\includegraphics[width=\linewidth]{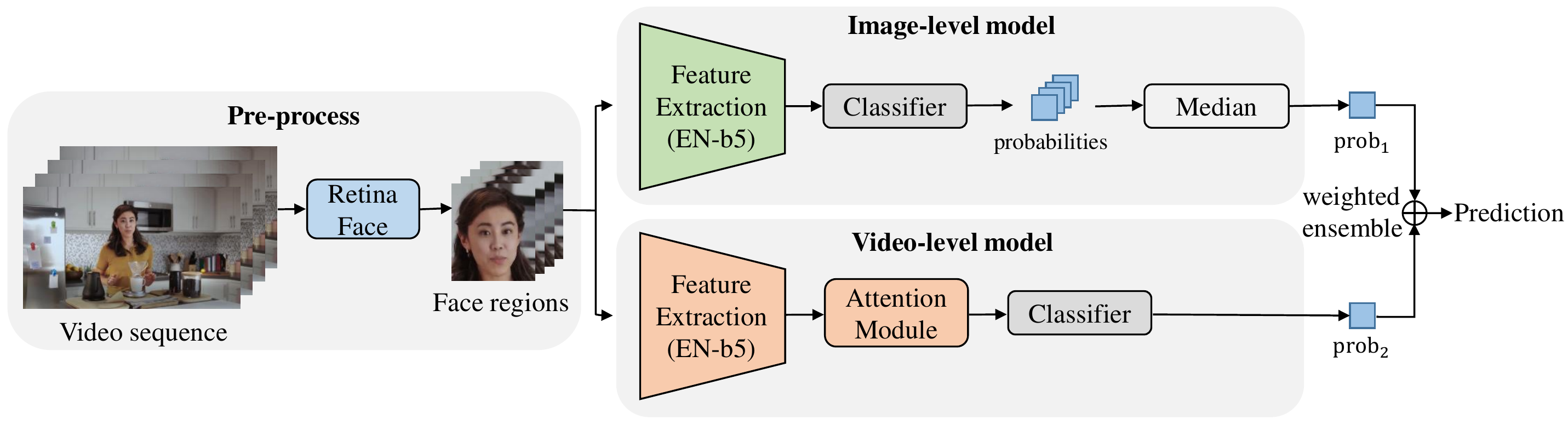}
%	\vspace{-0.5cm}
	\caption{The framework of the second-place solution.}
	\label{fig:solution2}
%	\vspace{-0.16cm}
\end{figure}

Face manipulated video contains two types of forgery traces, \ie, image-level artifacts and video-level artifacts.
The former refers to the artifacts such as blending boundaries and abnormal textures within image, while the latter is the face jitter problem between video frames.
Most previous works only focused on artifacts in a specific modality and lacked consideration of both.
The team in the second place proposes to use an attention mechanism to fuse the temporal information in videos, and further combine it with an image model to achieve better results.

The overall framework of their method is shown in Figure~\ref{fig:solution2}.
First, they use RetinaFace~\cite{retinaface} with $20\%$ margin to detect faces in video frames.
Then, the face sequence is fed into an image-based model and a video-based model, where the backbones are both EfficientNet-b5~\cite{efficientnet} with NoisyStudent~\cite{noisystudent} pre-trained weights.
The image-based model predicts frame by frame and takes the median of probabilities as the prediction.
The video-based model takes the entire face sequence as the input and adopts an attention module to fuse the temporal information between frames.
Finally, the per-video prediction score is obtained by averaging the probabilities predicted by the above two models.

\noindent
\textbf{Implementation Details.}
The team implements the proposed method via PyTorch. All the models are trained on $8$ NVIDIA Tesla V100 GPUs.
In addition to the DeeperForensics-1.0 dataset, they use three external datasets, \ie, FaceForensics++~\cite{FF++iccv}, Celeb-DF~\cite{celebdfcvpr}, and Diverse Fake Face Dataset~\cite{DFFD}.
They used the official splits provided by the above datasets to construct the training, validation and test sets. They balance the positive and negative samples through the down-sampling technique.

\noindent
\textit{$\bullet$ Training:}
The second-place team uses the following data augmentations: RandAugment~\cite{randaugment}, patch Gaussian~\cite{patchgaussian}, Gaussian blur, image compression, random flip, random crop and random brightness contrast.
They also employ the perturbation implementation in DeeperForensics-1.0~\cite{drf1_aug}.
For the image-based model, they train a classifier based on EfficientNet-b5~\cite{efficientnet}, using binary cross-entropy loss as the loss function.
They adopt a two-stage training strategy for the video-based model.
In stage-1, they train an image-based classifier based on EfficientNet-b5.
In stage-2, they fix the model parameters trained in stage-1 to serve as face feature extractor, and introduce an attention module to learn temporal information via nonlinear transformations and \textit{softmax} operations.
The input of the network is the face sequence (\ie, $5$ frames per video) in stage-2, and only the attention module and classification layers are trained. The binary cross-entropy loss is adopted as the loss function.
The input size is scaled to $320\times320$.
Adam optimizer~\cite{adam} is used with a learning rate of $0.0002$, $\beta_1=0.9$, $\beta_2=0.999$, and weight decay of $0.00001$. The batch size is $32$. 
The total number of training epochs is set to $20$, and the learning rate is halved every $5$ epochs.

\noindent
\textit{$\bullet$ Testing:}
They sample $10$ frames at equal intervals for each video and detect faces by RetinaFace~\cite{retinaface} as in the training phase.
Then, the face images are resized to $320\times320$. Test-time augmentation (TTA) (\eg, flip) is applied to get $20$ images ($10$ original and $10$ flipped), which are fed into network to get the prediction score.
They clip the prediction score of each video to $[0.01, 0.99]$ to avoid excessive losses on extreme error samples.

\subsection{Solution of Third Place}
\label{sec:solution3}

\noindent
\textit{Team members: Changlei Lu, Ganchao Tan}

\begin{figure}[h]
	\centering
%	\vspace{-0.35cm}
	\includegraphics[width=0.9\linewidth,height=0.577\linewidth]{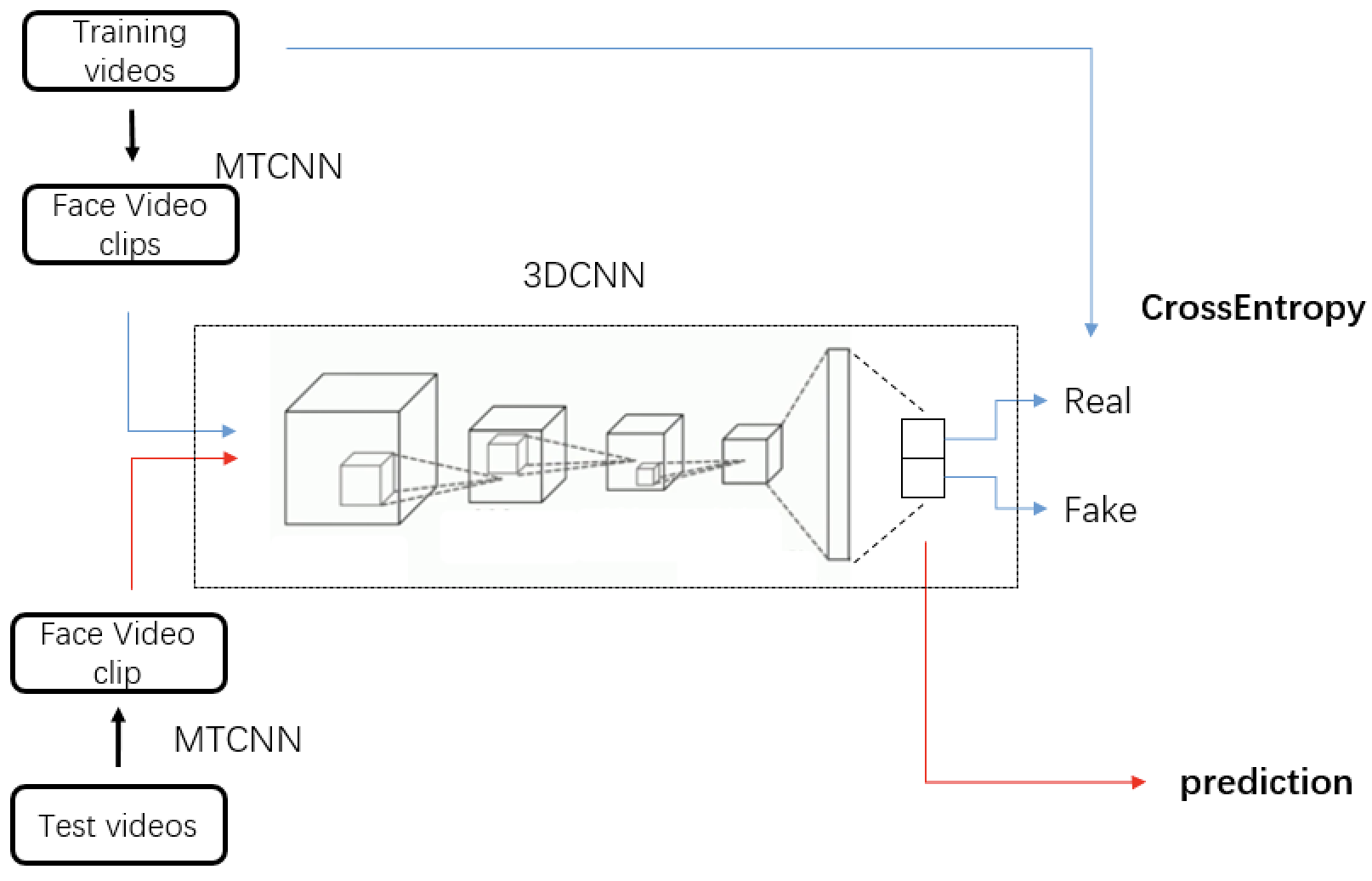}
%	\vspace{-0.5cm}
	\caption{The framework of the third-place solution.}
	\label{fig:solution3}
%	\vspace{-0.16cm}
\end{figure}

Similar to the second-place entry, the team in the third place also utilize the poor temporal consistency in existing face manipulation techniques.
%
%Thus, temporal information can be an important cue for real-world face forgery detection.
%
To this end, they propose to use a 3D convolutional neural network (3DCNN) to capture spatial-temporal features for forgery detection.
The framework of their method is shown in Figure~\ref{fig:solution3}.

\noindent
\textbf{Implementation Details.}
First, the team crops faces in the video frames using the MTCNN~\cite{mtcnn} face detector. They combine all the cropped face images into a face video clip.
Each video clip is then resized to $64\times224\times224$ or $64\times112\times112$.
Various data augmentations are applied, including Gaussian blur, white Gaussian noise in color components, random crop, random flip, \etc.
Then, they use the processed video clips as the input to train a 3D convolutional neural network (3DCNN) using the cross-entropy loss. They examine three kinds of networks, I3D~\cite{i3d}, 3D ResNet~\cite{3dresnet} and R(2+1)D~\cite{r2plus1d}. These models are pre-trained on the action recognition datasets, \eg, kinetics~\cite{kinetics}.
In addition to DeeperForensics-1.0, they use three external public face manipulation datasets, \ie, the DFDC dataset~\cite{DFDCChallenge}, Deep Fake Detection~\cite{google}, and FaceForensics++~\cite{FF++iccv}.

%% file: sections/discussion.tex
% !TEX root = ../main.tex

\section{Discussion}
\label{sec:discussion}

The methods mentioned above have considered different potential aspects in developing a robust face forgery detection model.
We are glad to find the winning solutions achieve promising performance in the DeeperForensics Challenge 2020.
In summary, there are three key points inspired by these methods that could improve real-world face forgery detection.
1) Strong backbone. Backbone selection of the forgery detection models is important. The high-performance winning solutions are based on state-of-the-art EfficientNet.
2) Diverse augmentations. Applying appropriate data augmentations may better simulate real-world scenarios and boost the model performance.
3) Temporal information. Since the primary detection target is the fake videos, temporal information can be a critical clue to distinguish the real from the fake.

 Despite the promising results, we believe that there is still much room for improvement in the real-world face forgery detection task.
 1) More suitable and diverse data augmentations may contribute to a better simulation of real-world data distribution.
 2) Developing a robust detection method that can cope with unseen manipulation methods and distortions is a critical problem. At this stage, we observe that the model training is data-dependent. Although data augmentations can help improve the performance to a certain extent, the generalization ability of most forgery detection models is still poor.
 3) Different artifacts in the Deepfakes videos (\eg, checkerboard artifacts, fusion boundary artifacts) remain rarely explored.

\vspace{0.1cm}
\noindent
\textbf{Acknowledgments.} We thank Amazon Web Services for sponsoring the prize of this challenge. The organization of this challenge is also supported by A*STAR through the Industry Alignment Fund - Industry Collaboration Projects Grant.